%% file: specificity_paper.tex
\documentclass[10pt,twocolumn,letterpaper]{article}

\usepackage{titling}
\usepackage{cvpr}
\usepackage{times}
\usepackage{epsfig}
\usepackage{graphicx}
\usepackage{amsmath}
\usepackage{amssymb}
\usepackage{multirow}

\usepackage{mdwlist}
\usepackage{mysymbols}
\usepackage{enumitem}
\usepackage{booktabs} %

\usepackage[pagebackref=true,breaklinks=true,letterpaper=true,colorlinks,bookmarks=false]{hyperref}

\usepackage{glossaries}
\makeglossaries

\newenvironment{dummy}{}

\newglossarystyle{mytree}{%
 \glossarystyle{tree}

 } 
 
  \newglossarystyle{mylist}{%
 \renewcommand*{\glsgroupheading}[1]{}%
 } 

\glossarystyle{mylist}

\newglossaryentry{human specificity}
{
  name=human specificity,
  description={Specificity measured from image textual descriptions by averaging human-annotated sentence similarities (Section~\ref{sec:human_spec})}
}
\newglossaryentry{automated specificity}
{
  name=automated specificity,
  description={Specificity computed from image textual descriptions by averaging automatically computed sentence similarities (Section ~\ref{sec:compute_similarity})}
}
\newglossaryentry{predicted specificity}
{
  name=predicted specificity,
  description={Specificity computed from image features without any textual descriptions}
}
\newglossaryentry{ground truth LR}
{
  name=ground-truth LR/specificity,
  description={Specificity computed using Logistic Regression parameters estimated from image textual descriptions (Section~\ref{sec:gt_spec})}
}
\newglossaryentry{predicted LR}
{
  name=predicted LR/specificity,
  description={Specificity computed using Logistic Regression parameters predicted from image features (Section~\ref{sec:predict_specificity})}
}

\usepackage[rgb]{xcolor}
\usepackage[author={Devi Parikh}, open=true]{pdfcomment}

\cvprfinalcopy %

\def\cvprPaperID{225} %

\begin{document}

\newcommand{\mainak}{\textcolor{magenta}}
\newcommand{\devi}{\textcolor{red}}
\newcommand{\mainakcomment}{\pdfcomment[author=Mainak Jas, color=yellow]}
\newcommand{\mainakst}{\pdfmarkupcomment[markup=StrikeOut,color=magenta]}

\title{Image Specificity}

\author{Mainak Jas\\
Aalto University\\
{\tt\small mainak.jas@aalto.fi}
\and
Devi Parikh\\
Virginia Tech\\
{\tt\small parikh@vt.edu}
}

\maketitle
\input{sections/abstract}
\input{sections/introduction}
\input{sections/related_work}
\input{sections/approach}
\input{sections/results}
\input{sections/discussion}

\paragraph{Acknowledgements:}This work was supported in part by The Paul G. Allen Family Foundation Allen Distinguished Investigator award  to D.P.

\newpage

\title{Image Specificity \\ (Supplementary material)}

\maketitle

\input{sections/supp}

\printglossary

{\small
\bibliographystyle{ieee}
\bibliography{sections/references}
}

\end{document}

%% file: sections/abstract.tex
\begin{abstract}

For some images, descriptions written by multiple people are consistent with each other. But for other images, descriptions across people vary considerably. In other words, some images are specific -- they elicit consistent descriptions from different people -- while other images are ambiguous. Applications involving images and text can benefit from an understanding of which images are specific and which ones are ambiguous. For instance, consider text-based image retrieval. If a query description is moderately similar to the caption (or reference description) of an ambiguous image, that query may be considered a decent match to the image. But if the image is very specific, a moderate similarity between the query and the reference description may not be sufficient to retrieve the image.

In this paper, we introduce the notion of \emph{image specificity}. We present two mechanisms to measure specificity given multiple descriptions of an image: an automated measure and a measure that relies on human judgement. We analyze image specificity with respect to image content and properties to better understand what makes an image specific. We then train models to automatically predict the specificity of an image from image features alone without requiring textual descriptions of the image. Finally, we show that modeling image specificity leads to improvements in a text-based image retrieval application.
\end{abstract}

%% file: sections/introduction.tex
\vspace{-10pt}
\section{Introduction}
\begin{figure}[t]
\begin{center}
   \includegraphics[width=1\linewidth]{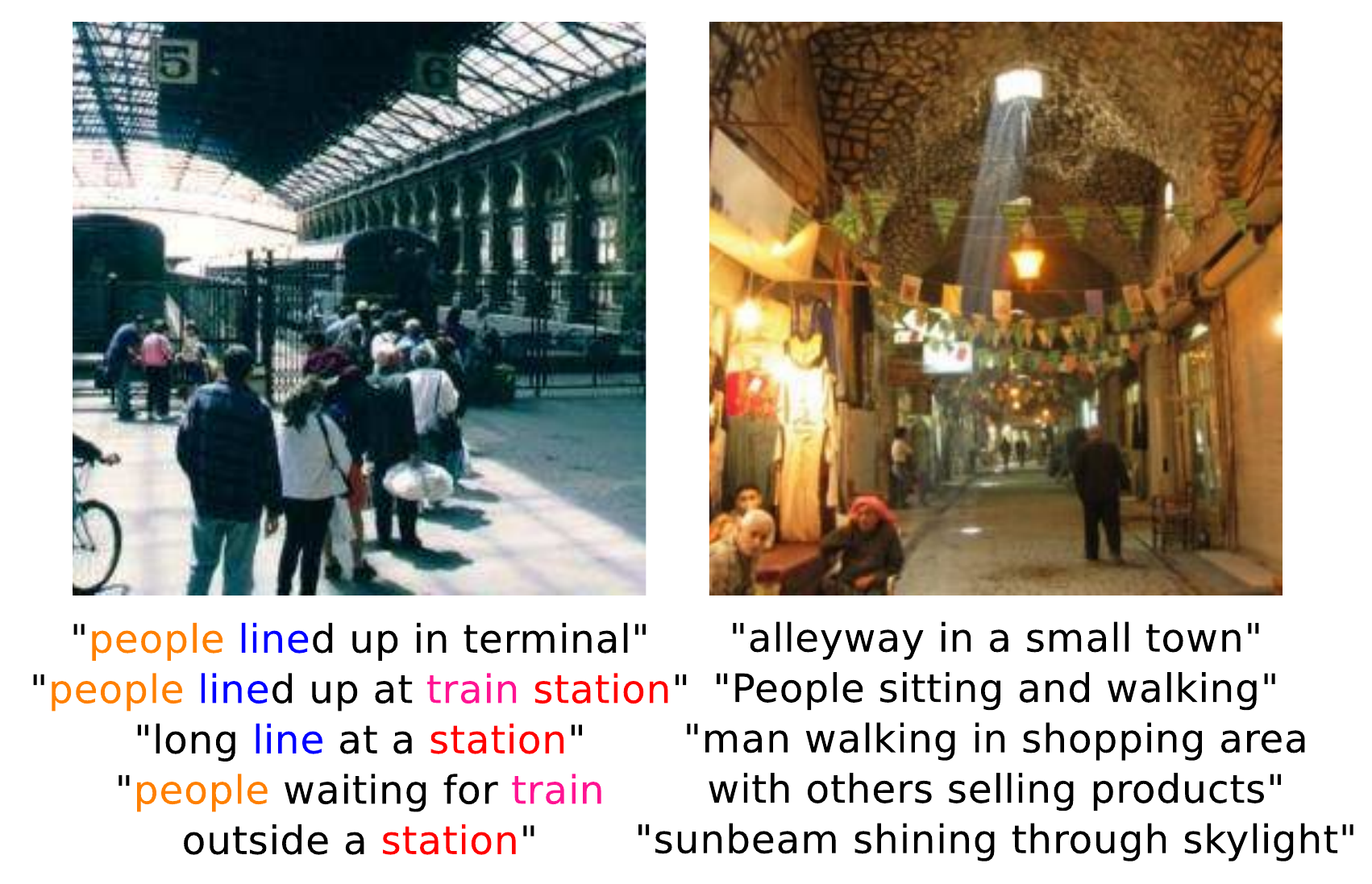}
\end{center}
\vspace{-12pt}
   \caption{
   Some images are \emph{specific} -- they elicit consistent descriptions from different people (left). Other images (right) are ambiguous.
   }
   \vspace{-10pt}
\label{fig:specificity_intro}
\end{figure}

Consider the two photographs in Figure~\ref{fig:specificity_intro}. How would you describe them? For the first, phrases like ``people lined up in  terminal'', ``people lined up at train station'', ``people waiting for train outside a station'', \etc. come to mind. It is clear what to focus on and describe. In fact, different people talk about similar aspects of the image -- the train, people, station or terminal, lining or queuing up. But for the photograph on the right, it is less clear how it should be described. Some people talk about the the sunbeam shining through the skylight, while others talk about the alleyway, or the people selling products and walking. %
In other words, the photograph on the left is \emph{specific} whereas the photograph on the right is \emph{ambiguous}. 

The computer vision community has made tremendous progress on recognition problems such as object detection~\cite{everinghampascal, girshick2013rich}, image classification~\cite{krizhevsky2012imagenet}, attribute classification~\cite{wang2009joint} and scene recognition~\cite{xiao2010sun, zhou2014places}.  Various approaches are moving to higher-level semantic image understanding tasks. One such task that is receiving increased attention in recent years is that of automatically generating textual descriptions of images~\cite{arxivpaper1,  arxivpaper6, arxivpaper2, farhadi2010every, arxivpaper3, arxivpaper4, kulkarni2011baby, li2011composing, arxivpaper7, mitchell2012midge, ordonez2011im2text, arxivpaper5, yao2010i2t} and evaluating these descriptions~\cite{elliott2014comparing, lin2004rouge, ordonez2011im2text, papineni2002bleu}. However, these works have largely ignored the \emph{variance in descriptions} produced by different people describing each image. %
In fact, early works that tackled the image description problem~\cite{farhadi2010every} or reasoned about what image content is important and frequently described~\cite{berg2012understanding} claimed that human descriptions are consistent. We show that there is in fact variance in how consistent multiple human-provided descriptions of the same image are. Instead of treating this variance as noise, we think of it as a useful signal that if modeled, can benefit applications involving images and text. %

We introduce the notion of \emph{image specificity} which measures the amount of variance in multiple viable descriptions of the same image. Modeling image specificity can benefit a variety of applications. For example, computer-generated image description and evaluation approaches can benefit from specificity. If an image is known to be ambiguous, several different descriptions can be generated and be considered to be plausible. But if an image is specific, a narrower range of descriptions may be appropriate. %
Photographers, editors, graphics designers, \etc. may want to pick specific images -- images that are likely to have a single (intended) interpretation across viewers. %

Given multiple human-generated descriptions of an image, we measure specificity using two different mechanisms: one requiring human judgement of similarities between two descriptions, and the other using an automatic textual similarity measure. Images with a high average similarity between pairs of sentences describing the image are considered to be specific, while those with a low average similarity are considered to be ambiguous. %
We then analyze the correlation between image specificity and image content or properties to understand what makes certain images more specific than others. We find that images with people tend to be specific, while mundane images of generic buildings or blue skies do not tend to be specific. We then train models that can predict the specificity of an image just by using image features (without  associated human-generated descriptions). Finally, we leverage image specificity to improve performance in a real-world application: text-based image retrieval.

%% file: sections/related_work.tex
\section{Related work}
\paragraph{Image properties:}

Several works study high-level image properties beyond those depicted in the image content itself. 
For instance, unusual photographs were found be interesting~\cite{gygli2013interestingness} and images of indoor scenes with people were found to be memorable, while scenic, outdoor scenes were not~\cite{IsolaParikhTorralbaOliva2011, Isola2011}. Other properties of images such as aesthetics~\cite{dhar2011high}, attractiveness~\cite{leyvand2008data}, popularity~\cite{khosla2014makes}, and visual clutter~\cite{rosenholtz2007measuring} have also been studied\footnote{Our work is complementary to visual metamers~\cite{freeman2011metamers}. In visual metamers, different images are perceived similarly but in specificity, we study how the same image can be perceived differently, and how this variance in perception differs across images.}. In this paper, we study a novel property of images -- specificity -- that captures the degree to which multiple human-generated descriptions of an image vary. We study what image content and properties make images specific. We go a step further and leverage this new property to improve a text-based image retrieval application.

\vspace{-15pt}
\paragraph{Importance:}
Some works have looked at what is worth describing in an image. Bottom-up saliency models~\cite{itti1998model, Judd_2009} study which image features predict eye fixations. Importance~\cite{berg2012understanding, spain2007measuring} characterizes the likelihood that an object in an image will be mentioned in its description. Attribute dominance~\cite{turakhia2013attribute} models have been used to predict which attributes pop out and the order in which they are likely to be named. However, unlike most of these works, we look at the \emph{variance} in human perception of what is worth mentioning in an image and how it is mentioned.

\vspace{-5pt}
\paragraph{Image description:}
Several approaches have been proposed for automatically describing images. This paper does not address the task of generating descriptions. Instead, it studies a property of how humans describe images -- some images elicit consistent descriptions from multiple people while others do not. This property can benefit image description approaches. Some image description approaches are data-driven. They retrieve images from a database that are similar to the input image, and leverage descriptions associated with the retrieved images to describe the input image~\cite{farhadi2010every, ordonez2011im2text}. In such approaches, knowledge of the specificity of the input image may help guide the range of the search for visually similar images. If the input image is specific, perhaps only highly similar images and their associated descriptions should be used to construct its description. Other approaches analyze the content of the image and then compose descriptive sentences using knowledge of sentence structures~\cite{kulkarni2011baby, mitchell2012midge}. For images that are ambiguous, the model can predict multiple diverse high-scoring descriptions of the image that can all be leveraged for a downstream application. Finally, existing automatic image description evaluation metrics such as METEOR~\cite{banerjee2005meteor}, ROUGE~\cite{lin2004rouge}, BLEU~\cite{papineni2002bleu} and CIDEr~\cite{1411.3041} compare a generated description with human-provided reference descriptions of the image. This evaluation protocol does not account for the fact that some images have multiple viable ways in which they can be described. Perhaps the penalty for not matching reference descriptions of ambiguous images should be less than for specific ones.

\vspace{-15pt}
\paragraph{Image retrieval:} Query- or text-based image and video retrieval approaches evaluate how well a query matches the  content of~\cite{barnard2001learning, douze2011combining, lin2014visual, siddiquie2011image} or captions (descriptions) associated with~\cite{barnard2001learning, lew2000next}  images in a database. However, the fact that each image may have a different match score or similarity that is sufficient to make it relevant to a query has not been studied. In this work, we use image specificity to fill this gap. While the role of lexical ambiguity in information retrieval has been studied before~\cite{krovetz1992lexical}, reasoning about inherent ambiguity in images for retrieval tasks has not been explored.

%% file: sections/approach.tex
\section{Approach}

We first describe the two ways in which we measure the specificity of an image. We then describe how we use specificity in a text-based image retrieval application.

\subsection{Measuring Specificity} \label{sec:annotate_specificity}

We define the specificity of an image as the average similarity between pairs of sentences describing the image. For each image $i$, we are given a set $S^i$ of $N$ sentence descriptions $\{s^i_1, \ldots, s^i_N\}$. We measure the similarity between all possible $\binom{N}{2}$ pairs of sentences and average the scores. The similarity between two sentences can either be judged by humans or computed automatically.

\begin{figure*}[hbt!]
\begin{center}
   \includegraphics[width=1.0\linewidth]{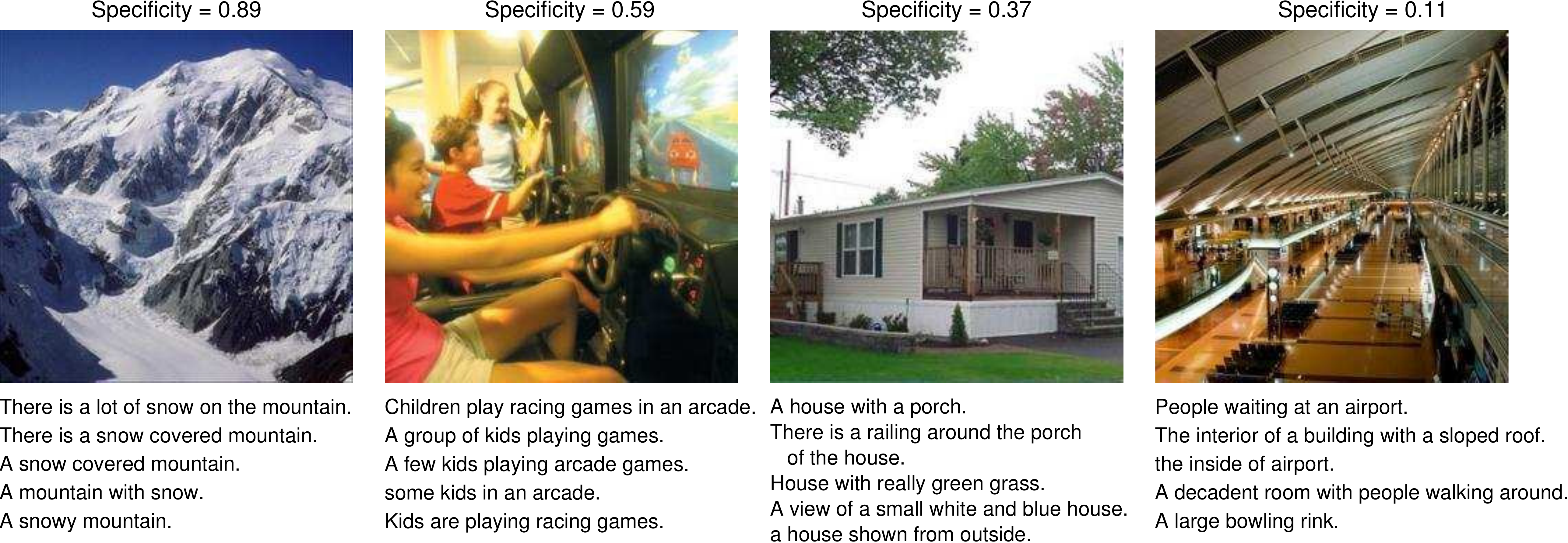}
\end{center}
\vspace{-12pt}
   \caption{Example images with very low to very high \glslink{human specificity}{human-annotated specificity} scores.}
   \vspace{-12pt}
\label{fig:specificity_examples}
\end{figure*}

\vspace{-7pt}
\subsubsection{Human Specificity Measurement}
\vspace{-7pt}
\label{sec:human_spec}
$M$ different subjects on Amazon Mechanical Turk (AMT) were asked to rate the similarity between a pair of sentences $s^i_a$ and $s^i_b$, on a scale of 1 (very different) to 10 (very similar). Note that subjects were not shown the corresponding image and were not informed that the sentences describe the same image. This ensured that subjects rated the similarity between sentences based solely on their textual content. We shift and scale the similarity scores to lie between 0 and 1. We denote this similarity, as assessed by the $m\text{-th}$ subject to be $\mathit{sim}_\mathrm{hum}^m(s^i_a, s^i_b)$

The average similarity score across all pairs of sentences and subjects gives us the specificity score $\mathit{spec}_\mathrm{hum}^i$ for image $i$ based on human perception. For ease of notation, we drop the superscript $i$ when it is clear from the context.

\vspace{-12pt}
\begin{equation} \label{eq:specificity}
\textit{spec}_{\glslink{human specificity}{\textrm{hum}}} = \frac{1}{M \binom{N}{2}} \sum_{\forall \{s_a,s_b\} \subset S} \sum_{m=1}^{M} \mathit{sim}_\mathrm{hum}^m(s_a, s_b)
\end{equation}

Figure~\ref{fig:specificity_examples} shows images with their \glslink{human specificity}{human-annotated specificity} scores. Note how the specificity score drops as the sentence descriptions become more varied.

\vspace{-10pt}
\subsubsection{Automated Specificity Measurement}
\vspace{-7pt}
\label{sec:compute_similarity}
To measure specificity automatically given the $N$ descriptions for image $i$, we first tokenize the sentences and only retain words of length three or more. This ensured that semantically irrelevant words, such as `a', `of', \etc, were not taken into account in the similarity computation (a standard stop word list could also be used instead). We identified the synsets (sets of synonyms that share a common meaning) to which each (tokenized) word belongs using the Natural Language Toolkit~\cite{bird2006nltk}. Words with multiple meanings can belong to more than one synset. Let $Y_{au} = \{y_{au}\}$ be the set of synsets associated with the $u\text{-th}$ word from sentence $s_a$. 

Every word in both sentences contributes to the automatically computed similarity $\mathit{sim}_\mathrm{auto}(s_a, s_b)$ between a pair of sentences $s_a$ and $s_b$. The contribution of the $u\text{-th}$ word from sentence $s_a$ to the similarity is $c_{au}$. This contribution is computed as the maximum similarity between this word, and all words in sentence $s_b$ (indexed by $v$). The similarity between two words is the maximum similarity between all pairs of synsets (or senses) to which the two words have been assigned. We take the maximum because a word is usually used in only one of its senses. Concretely,

\vspace{-12pt}
\begin{equation}
c_{au} = \max_{v} \max_{y_{au} \in Y_{au}} \max_{y_{bv} \in Y_{bv}} \mathit{sim}_\mathrm{sense} (y_{au},y_{bv})
\end{equation}

The similarity between senses $\mathit{sim}_\mathrm{sense} (y_{au},y_{bv})$ is the shortest path similarity between the two senses on WordNet~\cite{miller1995wordnet}. We can similarly define $c_{bv}$ to be the contribution of $v\text{-th}$ word from sentence $s_b$ to the similarity $\mathit{sim}_\text{auto}(s_a, s_b)$ between sentences $s_a$ and $s_b$.

The similarity between the two sentences is defined as the average contribution of all words in both sentences, weighted by the importance of each word. Let the importance of the $u\text{-th}$ word from sentence $s_a$ be $t_{au}$. This importance is computed using term frequency-inverse document frequency (TF-IDF) using the scikit-learn software package~\cite{pedregosa2011scikit}. Words that are rare in the corpus but occur frequently in a sentence contribute more to the similarity of that sentence with other sentences. So we have

\vspace{-12pt}
\begin{equation} \label{eq:automated_sent_sim}
\mathit{sim}_\mathrm{auto}(s_a, s_b) = \frac{\sum_{u} t_{au}c_{au} + \sum_{v} t_{bv}c_{bv}}{\sum_{u} t_{au} + \sum_{v} t_{bv}}
\end{equation}

The denominator in Equation~\ref{eq:automated_sent_sim} ensures that the similarity between two sentences is independent of sentence-length and is always between 0 and 1. Finally, the \gls{automated specificity} score $\mathit{spec}_\mathrm{auto}$ of an image $i$ is computed by averaging these similarity scores across all sentence pairs:

\vspace{-10pt}
\begin{equation} \label{eq:specificity}
\mathit{spec}_{\glslink{automated specificity}{\mathrm{auto}}} = \frac{1}{\binom{N}{2}} \sum_{\forall \{s_a,s_b\} \subset S} \mathit{sim}_\text{auto}(s_a, s_b)
\end{equation}

The reader is directed to the supplementary material~\cite{jas2015supp} for a pictorial explanation of \gls{automated specificity} computation.

\subsection{Application: Text-based image retrieval}\label{sec:image_retrieval}

We now describe how we use image specificity in a text-based image retrieval application.

\vspace{-12pt}
\subsubsection{Setup}
\vspace{-7pt}
There is a particular image the user is looking for from a database of images. We call this the target image. The user inputs a query sentence $q$ that describes the target image. Every image in the database is associated with a single reference description $r_{i}$ (not to be confused with the ``training" pool of sentences $S^i$ described in Section~\ref{sec:annotate_specificity} used to define the specificity of an image). This can be, for example, the caption in an online photo database such as Flickr. The goal is to sort the images in the database according to their relevance score $\mathit{rel}^i$ from most to least relevant, such that the target image has a low rank.

\vspace{-12pt}
\subsubsection{Baseline Approach}
\vspace{-7pt}
\label{sec:baseline}
The baseline approach automatically computes a similarity $\mathit{sim}_\mathrm{auto}(q,r_i)$ between $q$ and $r_i$ using Equation~\ref{eq:automated_sent_sim}. All images in the database are sorted in descending order using this similarity score. That is, 

\vspace{-7pt}
\begin{equation}
\label{eq:baseline}
\mathit{rel}_{\text{baseline}}^i = \mathit{sim}_\text{auto}(q,r_i).
\end{equation}

The image whose reference sentence has the highest similarity to the query sentence gets ranked first while the image whose reference sentence has the lowest similarity to the query sentence gets ranked last.

\vspace{-12pt}
\subsubsection{Proposed Approach}
\vspace{-7pt}
\label{sec:gt_spec}
In the proposed approach, instead of ranking just by similarity between the query sentence and reference descriptions in the database, we take into consideration the specificity of each image. The rationale is the following: a specific image should be ranked high only if the query description matches the reference description of that image well, because we know that sentences that describe this image tend to be very similar. For ambiguous images, on the other hand, even mediocre similarities between query and reference descriptions may be good enough. 

This suggests that instead of just sorting based on $\mathit{sim}_\mathrm{auto}(q,r_i)$, the similarity between the query description $q$ and the reference description $r_i$ of an image $i$ (which is what the baseline approach does as seen in Equation~\ref{eq:baseline}), we should model $P(\mathrm{match}|\mathit{sim}_\mathrm{auto}(q,r_i))$ which captures the probability that the query sentence matches the reference sentence \ie, the query sentence describes the image. We use Logistic Regression (LR) to model this.

\vspace{-12pt}
\begin{equation}\begin{aligned} \label{eq:gt-specificity}
	\mathit{rel}_{\glslink{ground truth LR}{\mathrm{gt-specificity}}}^i & =  P(\mathrm{match}|\mathit{sim}_\text{auto}(q,r_i)) \\ 
	& = \left\{\frac{1}{1 + e^{-\beta_{0}^{i} - \beta_{1}^{i}\mathit{sim}_\mathrm{auto}(q,r_i)}}\right\} \\
\end{aligned}\end{equation}

For each image in the database, we train the above LR model. Positives examples of this model are the similarity scores between pairs of sentences both describing the image $i$ taken from the set $S^i$ described in Section~\ref{sec:annotate_specificity}. Negative examples are similarity scores between pairs of sentences where one sentence describes the image $i$ but the other does not. If there are $N$ descriptions available for each image during training, we have $\binom{N}{2}$ positive examples (all pairs of $N$  sentences). We generate a similar number of negative examples by pairing each of the $N$ descriptions with $\ceil{\frac{N-1}{2}}$ descriptions from other images. $\ceil{.}$ is the ceiling function.

The parameters of this LR model, $\beta_{0}^{i}$ and $\beta_{1}^{i}$, inherently capture the specificity of the image. Note that a separate LR model is trained for each image to model the specificity for that image. After these models have been trained, given a new query description $q$, the similarity $\mathit{sim}_\mathrm{auto}(q,r_i)$ is computed with every reference description $r_i$ in the dataset. The trained LR for each image, i.e. the parameters $\beta_{0}^{i}$ and $\beta_{1}^{i}$, can be used to compute $P(\mathrm{match}|\mathit{sim}_\mathrm{auto}(q,r_i))$ for that image. All images can then be sorted by their corresponding $P(\mathrm{match}|\mathit{sim}_\mathrm{auto}(q,r_i))$ values. In our experiments, unless mentioned otherwise, the query and reference descriptions being used at test time were not part of the training set used to train the LRs.

\vspace{-12pt}
\subsubsection{Predicting Specificity of Images}\label{sec:predict_specificity}
\vspace{-7pt}

The above approach needs several sentences per image to obtain positive and negative examples to train the LR. But in realistic scenarios, it may not be viable to collect multiple sentences for every image in the database. Hence, we learn a mapping from the image features $\{\boldsymbol{x_{i}}\}$ to the LR parameters estimated using sentence pairs. We call these parameters \glslink{ground truth LR}{ground-truth LR} parameters. We train two separate $\nu$-Support Vector Regression (SVR) models (one for each $\beta$ term) with Radial Basis Function (RBF) kernel. The learnt SVR model is then used to predict the LR parameters $\hat{\beta}_{0}^{i}$ and $\hat{\beta}_{1}^{i}$ of any previously unseen image. Finally, these \glslink{predicted LR}{predicted LR} parameters are used to compute $\hat{P}(\mathrm{match}|\mathit{sim}_\mathrm{auto}(q,r_i))$ and sort images in a database according to their relevance to a query description $q$. Of course, each image in the database still needs a (single) reference description $r_i$ (without which text-based image retrieval is not feasible).

\vspace{-12pt}
\begin{equation}\begin{aligned} \label{eq:pred-specificity}
	\mathit{rel}_{\glslink{predicted LR}{\mathrm{pred-specificity}}}^i & =  \hat{P}(\mathrm{match}|\mathit{sim}_\mathrm{auto}(q,r_i)) \\ 
	& = \left\{\frac{1}{1 + e^{-\hat{\beta}_{0}^{i} - \hat{\beta}_{1}^{i}\mathit{sim}_\mathrm{auto}(q,r_i)}}\right\} \\
\end{aligned}\end{equation}

Notice that the baseline approach (Equation~\ref{eq:baseline}) is a special case of our proposed approaches (Equations~\ref{eq:gt-specificity} and \ref{eq:pred-specificity}) with  $\beta^i_0 = \hat{\beta}^i_0 = \mathrm{constant}_0$ and $\beta^i_1 = \hat{\beta}^i_1 = \mathrm{constant}_1~\forall i$, where the parameters for each image are the same.

\vspace{-10pt}
\subsubsection{Summary} 
\vspace{-7pt}

Let's say we are given a new database of images and associated (single) reference descriptions that we want to search using query sentences. SVRs are used to predict each of the two LR parameters using image features for every image in the database. This is done offline. When a query is issued,  its similarity is computed to each reference description in the image. Each of these similarities are substituted into Equation~\ref{eq:pred-specificity} to calculate the relevance of each image using the LR parameters predicted for that image. This query-time processing is computationally light. The images are then sorted by the probability outputs of their LR models. The quality of the retrieved results using this (proposed) approach is compared to the baseline approach that sorts all images based on the similarity between the query sentence and reference descriptions. Of course, in the scenario where multiple reference descriptions are available for each image in the database, we can directly estimate the \glslink{ground truth LR}{(ground-truth) LR} parameters using those descriptions (as described in Section~\ref{sec:gt_spec}) instead of using the SVR to predict the LR parameters. We will show results of both approaches (using \glslink{ground truth LR}{ground-truth LR} parameters and \glslink{predicted LR}{predicted LR} parameters). 

%% file: sections/results.tex
\section{Experimental Results}

\subsection{Datasets and Image Features} %

We experiment with three datasets. The first is the MEM-5S dataset containing 888 images from the memorability dataset~\cite{Isola2011}, which are uniformly spaced in terms of their memorability. For each of these images, we collected 5 sentence descriptions by asking unique subjects on AMT to describe them.  Figure~\ref{fig:specificity_examples} shows some example images and their descriptions taken from the MEM-5S dataset. Since specificity measures the variance between sentences, and more sentences would result in a better specificity estimate, we also experiment with two datasets with 50 sentences per image in each dataset. One of these is the ABSTRACT-50S dataset~\cite{1411.3041} which is a subset of 500 images made of clip art objects from the Abstract Scene dataset~\cite{zitnick2013bringing} containing 50 sentences/image (48 training, 2 test). We use only the training sentences from this dataset for our experiments. The second is the PASCAL-50S dataset~\cite{1411.3041} containing 50 sentences/image for the 1000 images from the UIUC PASCAL dataset~\cite{rashtchian2010collecting}. These datasets allow us to study specificity in a wide range of settings, from real images to non-photorealistic but semantically rich abstract scenes. All correlation analysis reported in the following sections was performed using Spearman's rank correlation coefficient.

For predicting \glslink{predicted LR}{specificity}, we extract 4096D DECAF-6~\cite{donahue2013decaf} features from the PASCAL-50S images. Images in the ABSTRACT-50S dataset are represented by the occurrence, location, depth, flip angle of objects, object co-occurrences and clip art category (451D)~\cite{zitnick2013bringing}.

\subsection{Consistency analysis}

In Section~\ref{sec:annotate_specificity}, we described two methods to measure specificity. In the first, humans are involved in annotating the similarity score between the sentences describing an image and in the second, this is done automatically. We first analyze if humans agree on their notions of specificity, and then study how well \glslink{human specificity}{human annotation of specificity} correlates with \glslink{automated specificity}{automatically-computed specificity}. 

\vspace{-15pt}
\paragraph{Do humans rate sentence pair similarities consistently?}

The similarity of each pair of sentences in the MEM-5S dataset was rated by 3 subjects on AMT. This means that every image is annotated by $\binom{5}{2}*3=30$ similarity ratings. The average of the similarity ratings gives us the specificity scores. These ratings were split thrice into two parts such that the ratings from one subject was in one part and ratings from the other two subjects were in the other part. The specificity score computed from the first part was correlated with the specificity score of the other part. This gave an average correlation coefficient of $0.72$, indicating high consistency in specificity measured across subjects.

\vspace{-15pt}
\paragraph{Is specificity consistent for a new set of descriptions?}

Additionally, 5 more sentences were collected for a subset of 222 images in the memorability dataset. With these additional sentences, specificity was computed using human annotations and the correlation with the specificity from the previous set of sentences was found to be 0.54. Inter-human agreement on the same set of 5 descriptions for 222 images was 0.76. We see that specificity measured across two sets of five descriptions each is not highly consistent. Hence, we hypothesize that measuring specificity using more sentences would be desirable (thus our use of the PASCAL-50S and ABSTRACT-50S datasets)\footnote{It was prohibitively expensive to measure human specificity for all pairs of 50 sentences to verify this hypothesis}. %

\vspace{-15pt}
\paragraph{How do human and automated specificity compare?}

We find that the rank correlation between \glslink{human specificity}{human-annotated} and \glslink{automated specificity}{automatically measured specificity} (on the same set of 5 sentences for 888 images in MEM-5S) is 0.69 which is very close to the inter-human correlation of 0.72. Note that this automated method still requires textual descriptions by humans. In a later section, we will consider the problem of predicting specificity just from image features if textual descriptions are also not available. Note that in most realistic applications (\eg image search, that we explore later), it is practical to measure specificity by comparing descriptions automatically. Hence, the \gls{automated specificity} measurement may be the more relevant one.

\vspace{-15pt}
\paragraph{Are some images more specific than others?}

\begin{figure}[t]
\begin{center}
   \includegraphics[width=1.0\linewidth]{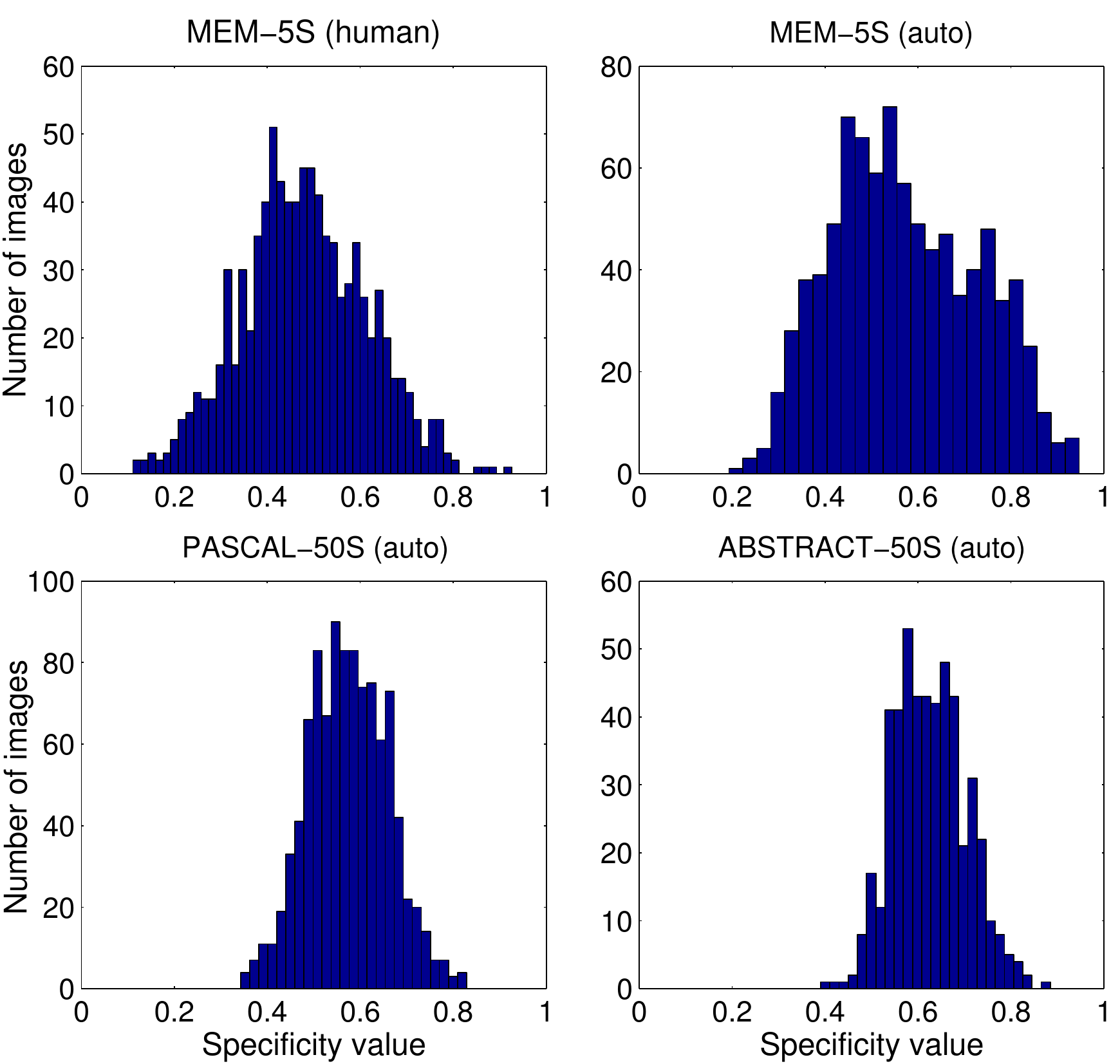}
\end{center}
\vspace{-12pt}
   \caption{A histogram of \glslink{human specificity}{human-annotated specificity} values for the MEM-5S dataset (top left) and \glslink{automated specificity}{automated specificity} values for all three datasets (rest).}
   \vspace{-12pt}
\label{fig:specificity_histogram}
\end{figure}

Now that we know specificity is well-defined, we study whether some images are in fact more specific than others. Figure~\ref{fig:specificity_examples} shows some examples of images whose specificity values range from low to very high values. Note how the descriptions become more varied as the specificity value drops. Figure~\ref{fig:specificity_histogram} shows a histogram of specificity values on all three datasets. In the MEM-5S dataset, the specificity values range from 0.11 to 0.93\footnote{Specificity values can fall between 0 and 1.}. This indicates that indeed, some images are specific and some images are ambiguous. We can exploit this fact to improve applications such as text-based image retrieval (Section~\ref{sec:image_search}). %

\subsection{What makes an image specific?}

We now study what makes an image specific. The first question we want to answer is whether longer sentence descriptions lead to more variability and hence less specific images. We correlated the average length of a sentence (measured as the number of words in the sentence) with specificity, and surprisingly, found that the length of a sentence had no effect on specificity ($\rho$=-0.02, p-value=0.64). However, we did find that the more specific an image was, the less was the variation in length of sentences describing it ($\rho$=--0.16, p-value$<$0.01).

\begin{figure}[t]
\begin{center}
   \includegraphics[width=0.9\linewidth]{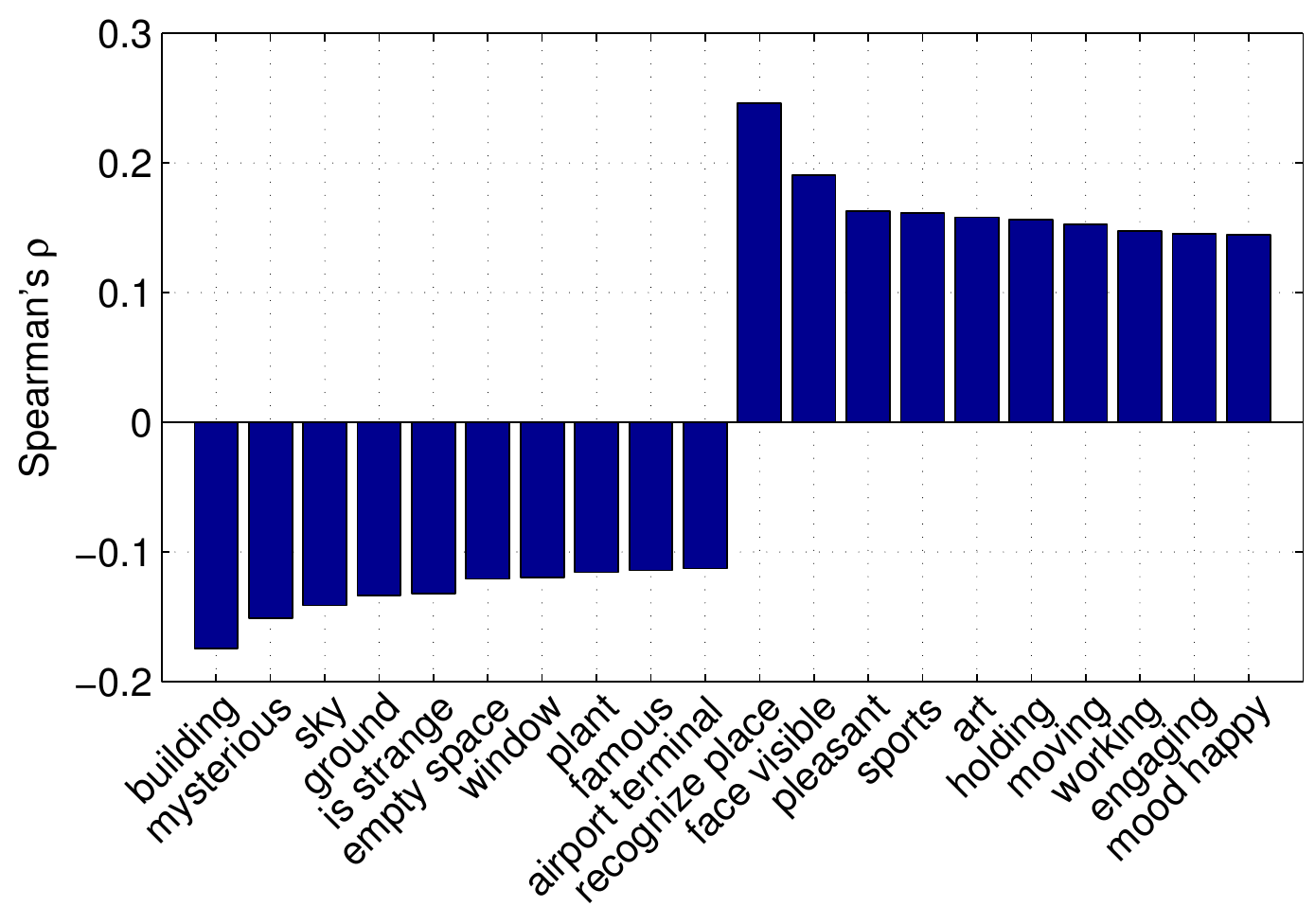}
\end{center}
\vspace{-12pt}
   \caption{Spearman's rank correlation of \gls{human specificity} with attributes, objects and scene annotations for the MEM-5S dataset.}
   \vspace{-12pt}
\label{fig:attribute_correlations}
\end{figure}

Next, we looked at image content to unearth possibly consistent patterns that make an image specific. We correlated publicly available attribute, object and scene annotations~\cite{IsolaParikhTorralbaOliva2011} for the MEM-5S dataset with our specificity scores. We then sorted the annotations by their correlation with specificity and showed the top 10 and bottom 10 correlations as a bar plot in Figure~\ref{fig:attribute_correlations}. We find that images with people tend to be specific, while mundane images of generic buildings or blue skies tend to not be specific. Note that if a category (\eg person) and its subcategory (\eg caucasian person) both appeared in the top 10 or bottom 10 list and had very similar correlations, the subcategory was excluded in favour of the main category since the subcategory is redundant. %

Next, we hypothesized that images with larger objects in them may be more specific, since different people may all talk about those objects. Confirming this hypothesis, we found a correlation of 0.16 with median object area and 0.14 with mean object area. %

We then investigated how importance~\cite{berg2012understanding} relates to specificity. Since important objects in images are those that tend to be mentioned often,  perhaps an image containing an important object will be more specific because most people will talk about the object. We consider all sentences corresponding to all images containing a certain object category.  In each sentence, we identify the word (\eg vehicle) that best matches the category (\eg car) using the shortest-path similarity of the words taken from the WordNet database~\cite{miller1995wordnet}. We average the similarity between this best matching word in each sentence to the category name across all sentences of all images containing that category. This is a proxy for how frequently that category is mentioned in descriptions of images containing the category. A similar average was found for randomly chosen sentences from other categories as a proxy for how often a category gets mentioned in sentences \textit{a priori}. These two averages were subtracted to obtain an importance value for each category. Now, the specificity scores of all images containing an object category was averaged and this score was found to be significantly correlated ($\rho$=0.31, p-value=0.05) with the importance score. This analysis was done only for categories that were present in more than 10 images in the MEM-5S dataset. This shows that images containing important objects do tend to be more specific.

In another study, Isola~\etal~\cite{Isola2011} measured image memorability. Images were flashed in front of subjects who were asked to press a button each time they saw the same image again. Interestingly, repeats of some images are more reliably detected across subjects than other images. That is, some images are more memorable than others. We tested if memorability and specificity are related by correlating them and found a high correlation ($\rho$=0.33, p-value$<$0.01) between the two. Thus, specificity can explain memorability to some extent. %
However, the two concepts are distinct. For instance, peaceful, picture-perfect scenes that may appear on a postcard or in a painting were found to be negatively correlated with memorability~\cite{IsolaParikhTorralbaOliva2011} ($\rho_\mathrm{peaceful}$=--0.35, $\rho_\mathrm{postcard}$=--0.31, $\rho_\mathrm{painting}$=--0.32). But these attributes have no correlation with specificity ($\rho_\mathrm{peaceful}$=--0.05, $\rho_\mathrm{painting}$=--0.02, $\rho_\mathrm{postcard}$=--0.05). In the supplementary material~\cite{jas2015supp}, we include examples of images that are memorable but not specific and vice-versa. Additional scatter plots for a subset of the computed correlations are also included in the supplementary material~\cite{jas2015supp}. Finally, correlation of mean color of the image with specificity was $\rho_{red}$=0.01, $\rho_{green}$=0.02 and $\rho_{blue}$=0.01. 

Overall, specificity is correlated with image content to quite an extent. In fact, if we train a regressor to predict \gls{automated specificity} directly from DECAF-6 features in the PASCAL-50S and MEM-5S dataset, we get a correlation of 0.2 and 0.25. The  correlation using semantic features in the ABSTRACT-50S dataset was 0.35. More details in supplementary material~\cite{jas2015supp}. The reader is  encouraged to browse our datasets through the websites on the authors' webpages.
\subsection{Image search} \label{sec:image_search}

\begin{figure}[t]
\begin{center}
   \includegraphics[width=0.85\linewidth]{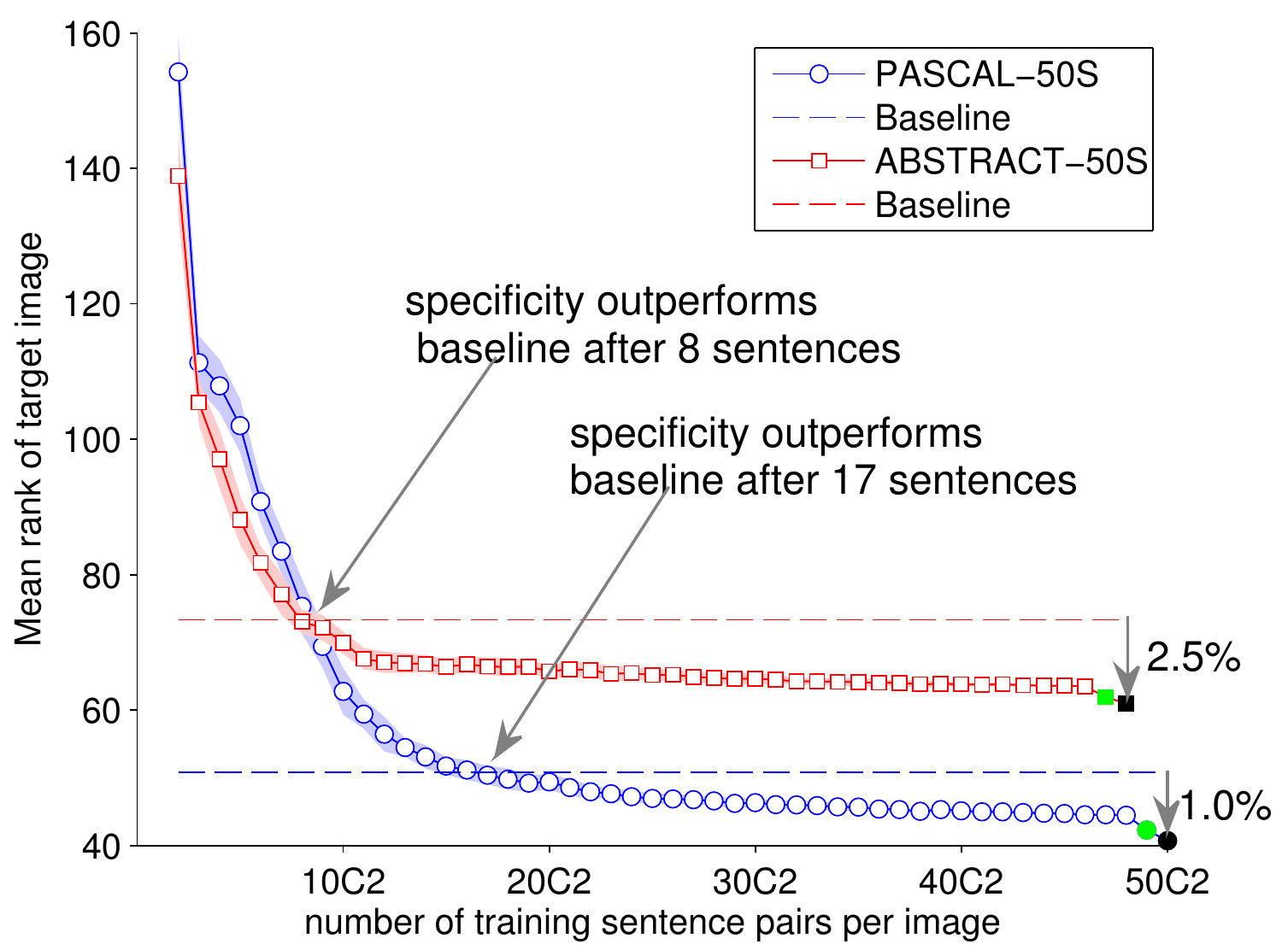}
\end{center}
\vspace{-12pt}
   \caption{Image search results: Increasing the number of training sentences per image improves the mean target rank obtained with \glslink{ground truth LR}{ground truth LR parameters} (specificity). As expected, there is a sharp improvement when the reference sentence (green fill) or both the reference and query sentences (black fill) are included when estimating the LR parameters. The results are averaged across 25 random repeats and the error intervals are shown in shaded colors. Annotations indicate the number of sentences required to beat baseline and the maximum improvement possible over baseline using all available sentences. Lower mean rank of target means better performance.}
   \vspace{-12pt}
\label{fig:training_sentences}
\end{figure}

\subsubsection{\glslink{ground truth LR}{Ground truth Specificity}}
\vspace{-7pt}
Given a database of images with multiple descriptions each, Section~\ref{sec:gt_spec} describes how we estimate parameters of a Logistic Regression (LR) model, and use the model for image search.
In our experiments, the query sentence corresponds to a known target image (known for evaluation, not to the search algorithm). The evaluation metric is the rank of the target images, averaged across multiple queries. 

We investigate the effect of number of training sentences per image used to train the LR model on the average rank of target images. Figure~\ref{fig:training_sentences} shows that the mean rank of the target image decreases with increasing number of training sentences. The baseline approach (Section~\ref{sec:baseline}) simply sorts the images by the similarity value between the query sentence and all reference sentences in the database (one per image). For the PASCAL-50S dataset, 17 training sentences per image were required to estimate an LR model that can beat this baseline while for ABSTRACT-50S dataset, 8 training sentences per image were enough. The improvement obtained over the baseline by training on all 50 sentences was 1.0\% of the total dataset size for PASCAL-50S and 2.5\% for the ABSTRACT-50S dataset\footnote{Our approach is general and can be applied to different automatic text similarity metrics. For instance, cosine similarity (dot product of the TF-IDF vectors) also works quite well with a 3.5\% improvement using \glslink{ground truth LR}{ground-truth specificity} for ABSTRACT-50S.}. With this improvement, we bridge 20.5\% of the gap between baseline and perfect result (target rank of 1) for PASCAL-50S and 17.5\% for ABSTRACT-50S.

\vspace{-10pt}
\subsubsection{\glslink{predicted LR}{Predicted Specificity}}
\vspace{-7pt}

\begin{table}[t]
\small
\begin{center}
\begin{tabular}{cccc}
\toprule
& Method & Mean & \% of queries \\
&        & rank & meet or beat BL \\
\midrule
\multirow{3}{*}{PASCAL-50S} & BL & 50.80 & --\\
& GT-Spec & 44.70 & 67.3\\
& P-Spec  & 49.72 & 73.2\\
\hline
\multirow{3}{*}{ABSTRACT-50S} & BL & 73.34 & --\\
& GT-Spec & 63.30 & 61.0\\
& P-Spec  & 69.41 & 61.6\\
\bottomrule
\end{tabular}
\end{center}
\vspace{-5pt}
\caption{Image search results for different ranking algorithms: baseline (BL), \glslink{ground truth LR}{specificity using ground truth (GT-Spec) LR parameters}, and \glslink{predicted LR}{specificity (P-Spec) using predicted LR parameters}. The column with header Mean rank gives the rank of the target image averaged across all images in the database. The final column indicates the percentage of queries where the method does better than or as good as baseline.}
\vspace{-12pt}
\label{table:statistics}
\end{table}

We noted in the previous section that as many as 17 training sentences per image are needed to estimate specificity accurately enough to beat baseline on the PASCAL-50S dataset. In a real application, it is not feasible to expect these many sentences per image. This leads us to explore if it is possible to predict specificity directly from images accurately enough to beat the baseline approach. 

As described in Section~\ref{sec:predict_specificity}, we train regressors that map image features to the LR parameters. These regressors are then used to predict the LR parameters that are used for ranking the database of test images in an image search application. We do this with leave-one-out cross-validation (ensuring that none of the textual descriptions of the predicted image were included in the training set) so that we have \glslink{predicted LR}{predicted LR parameters} on our entire dataset. %

\begin{figure}[t]
\begin{center}
   \includegraphics[width=0.8\linewidth]{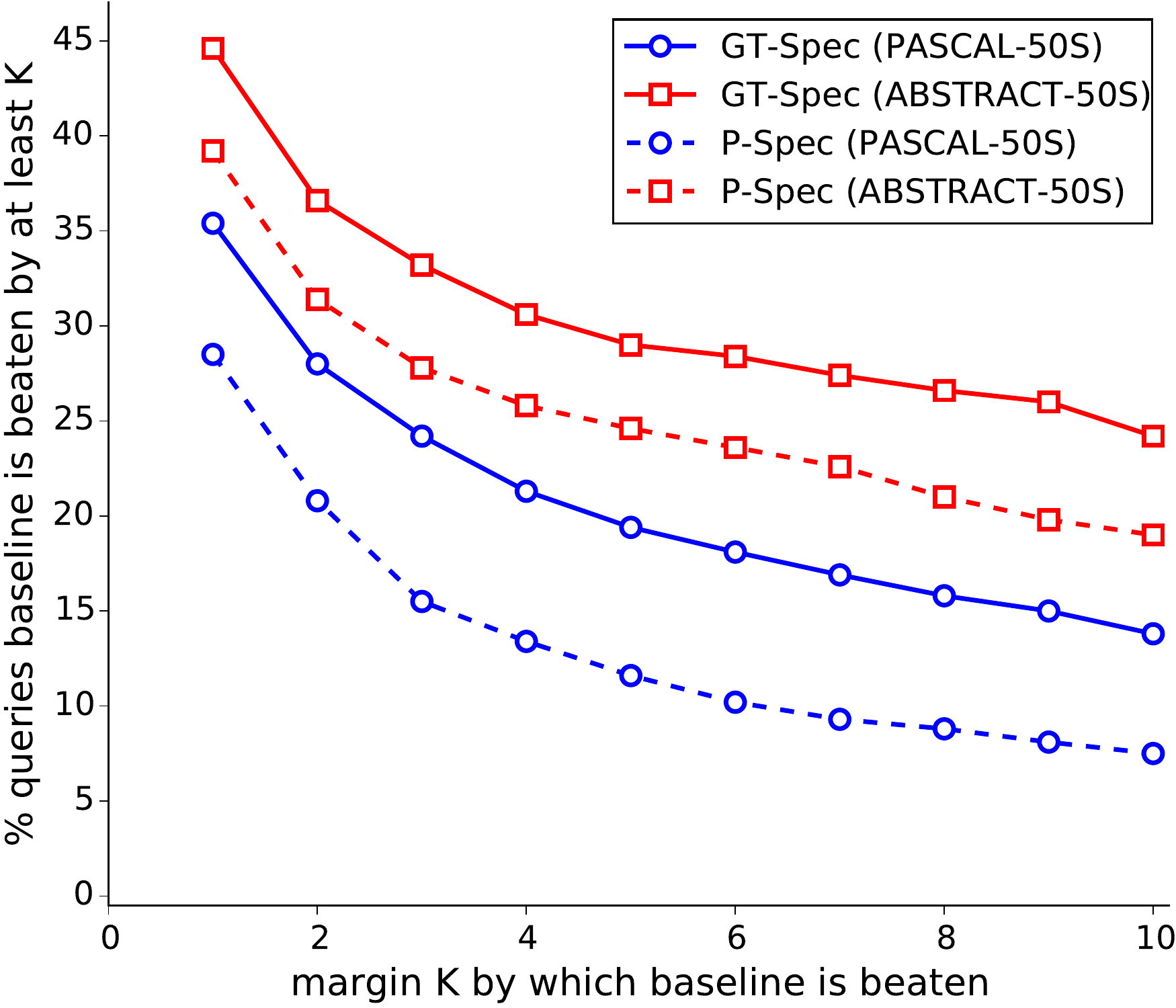}
\end{center}
\vspace{-12pt}
   \caption{Image search results: On the x-axis is plotted K, the margin in rank of target image by which baseline is beaten, and on the y-axis is the percentage of queries where baseline is beaten by at least K.}
   \vspace{-12pt}
\label{fig:retrieval_curve}
\end{figure}

Table~\ref{table:statistics} shows how often our approach does better than or matches the baseline.  It can be seen that specificity using the LR parameters predicted directly using image features does better than or matches the baseline for 73.2\% of the queries. This is especially noteworthy since \emph{no sentence descriptions} were used to estimate the specificity of the images. Specificity is predicted using purely image features.

From Table~\ref{table:statistics}, we note that \glslink{predicted LR}{predicted specificity} (P-Spec) loses less often to baseline as compared to \glslink{ground truth LR}{ground-truth specificity} (GT-Spec), but GT-Spec still has a better average rank of target images compared to P-Spec. The reason is that GT-Spec does \emph{much} better than P-Spec on queries where it wins against baseline. Therefore, we would like to know that when an approach beats baseline, how often does it beat baseline by a low or high margin? Figure~\ref{fig:retrieval_curve} shows the percentage of queries which beat baseline by different margins. The x-axis is the margin by at least which the baseline is beaten and on the y-axis is the percentage of queries. As expected, \glslink{ground truth LR}{ground-truth specificity} performs the best amongst the three methods. But even \glslink{predicted LR}{predicted specificity} often beats the baseline by large margins.

Many approaches~\cite{cui2008real, wang2011query} retrieve images based on text-based matches between the query and reference sentences, and then re-rank the results based on image content. This content-based re-ranking step can be performed on top of our retrieved results as well. Note that in our approach, the image features are used to modulate the similarity between the query and reference sentences -- and not to better assess the match between the query sentence and image content. These are two orthogonal sources of information.

Finally, Figure~\ref{fig:qualitative_search} shows a qualitative example from the ABSTRACT-50S dataset. In the first image, the query and the reference sentence for the target image do not match very closely. However, since the image has a low \glslink{automated specificity}{automated specificity}, this mediocre similarity is sufficient to lower the rank of the target image. %

%% file: sections/discussion.tex
\section{Discussion and Conclusion}

We introduce the notion of specificity. We present evidence that the variance in textual descriptions of an image, which we call specificity, is a well-defined phenomenon. We find that even abstract scenes which are not photorealistic capture this variation in textual descriptions. We study various object and attribute-level properties that influence specificity. 
More importantly, modeling specificity can benefit various applications. We demonstrate this empirically on a text-based image retrieval task. We use image features to predict the \emph{parameters} of a classifier (Logistic Regression) that modulates the similarity between the query and reference sentence differently for each image.
 Future work involves exploring robust measures of specificity that consider only representative sentences (not outliers), investigating other similarity measures such as Lin's similarity~\cite{lin1998information} or word2vec\footnote{\url{http://code.google.com/p/word2vec/}} when measuring specificity, exploring the potential of low-level saliency and objectness maps in predicting specificity, studying specificity in more controlled settings involving a closed set of visual concepts and using image specificity in various applications such as image tagging to determine how many tags to associate with an image (few for specific images and many for ambiguous images), image captioning, \etc Our data and code are publicly available on the authors' webpages. %
\begin{figure}[t]
\begin{center}
   \includegraphics[width=1.0\linewidth]{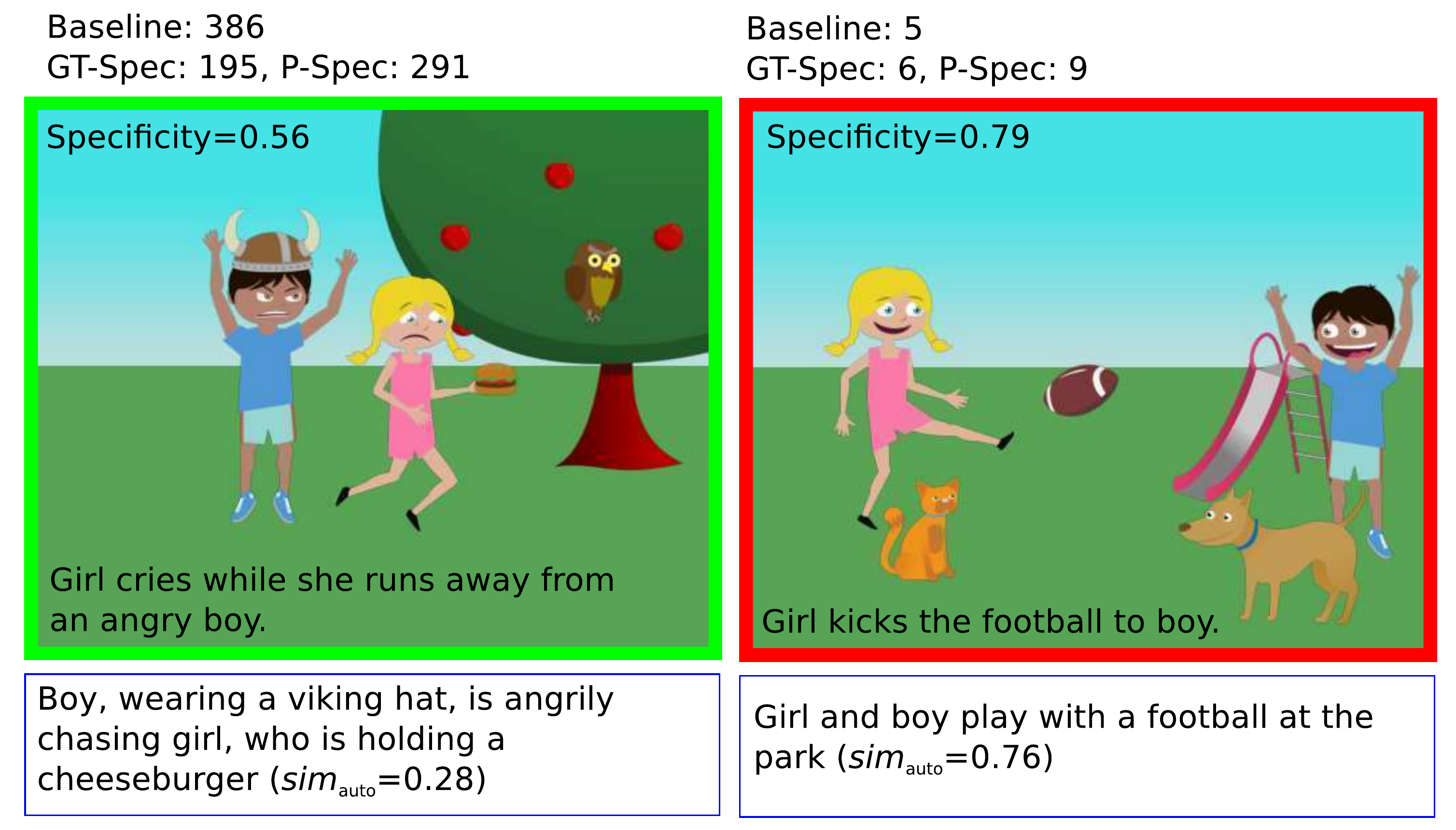}
\end{center}
\vspace{-12pt}
   \caption{Qualitative image search results from the ABSTRACT-50S dataset. The images are the target images. Query sentences are shown in the blue box below each image (along with their automated similarity to the reference sentence). The reference sentence of the image is shown on the image. The \gls{automated specificity} value is indicated at the top-left corner of the images. Green border (left) indicates that both \glslink{predicted LR}{predicted} and \glslink{ground truth LR}{ground-truth specificity} performed better than baseline, and red border (right) indicates that baseline did better than both P-Spec and GT-Spec. The rank of the target image using baseline, GT-Spec and P-Spec is shown above the image. %
   }
   \vspace{-12pt}
\label{fig:qualitative_search}
\end{figure}

%% file: sections/supp.tex
\section{Scatter plots for correlations}

\begin{figure}[htb!]
\begin{center}
   \includegraphics[width=0.8\linewidth]{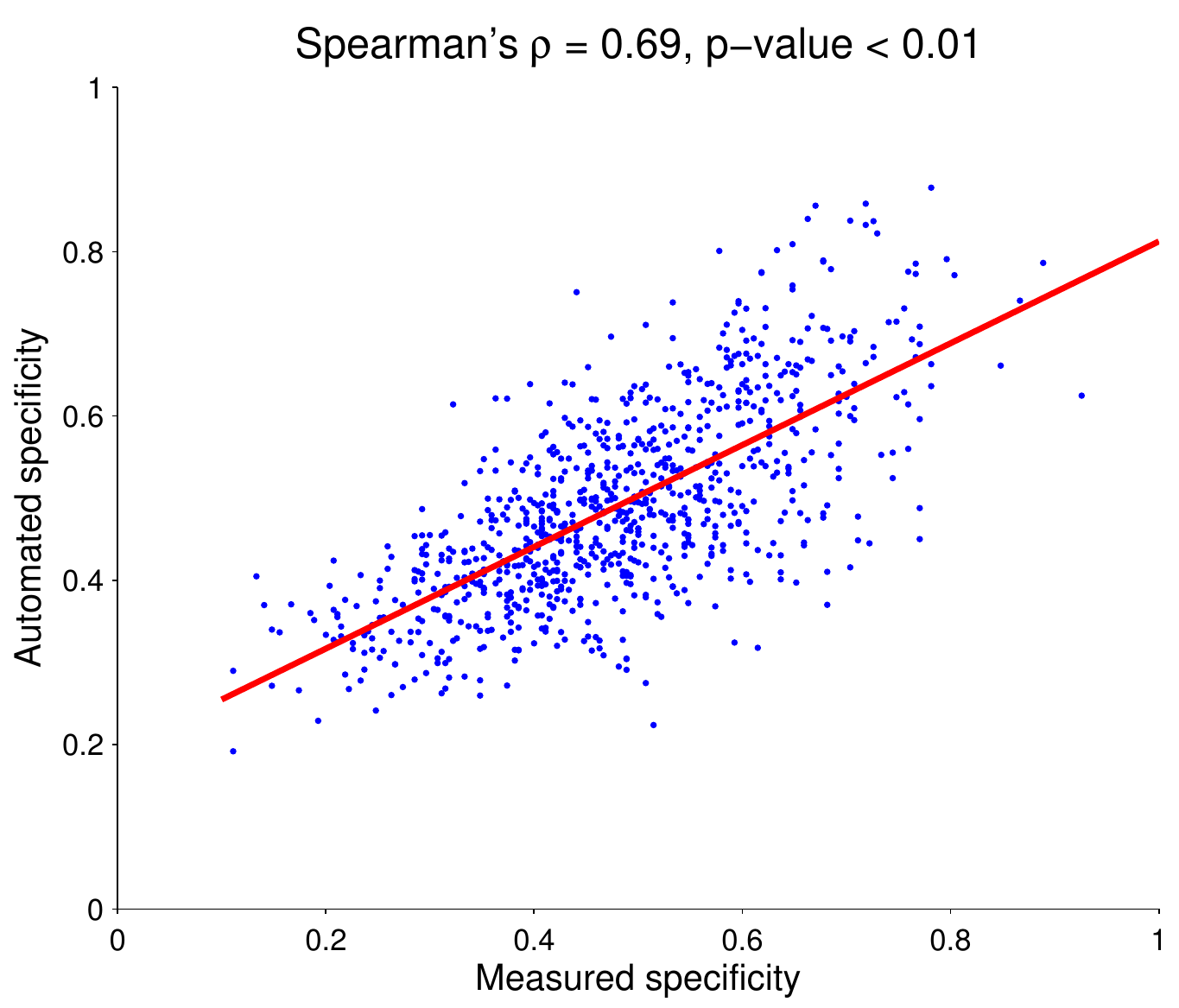}
\end{center}
   \caption{Correlation between \glslink{human specificity}{human-measured specificity} and \gls{automated specificity} for the MEM-5S dataset.}
\label{fig:automated_vs_human}
\end{figure}

\begin{figure}[htb]
\begin{center}
   \includegraphics[width=1.0\linewidth]{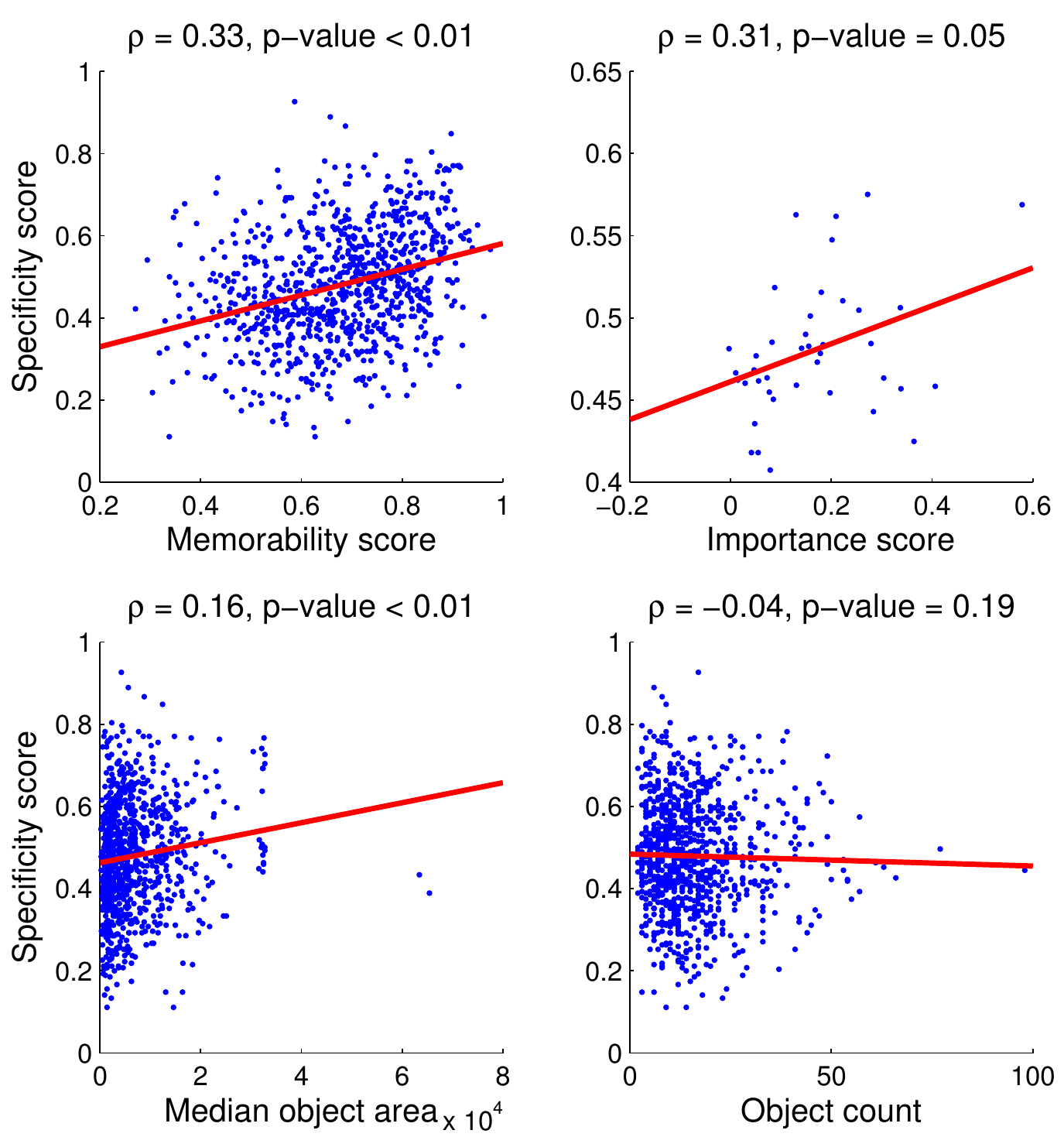}
\end{center}
   \caption{What makes an image specific? Memorable images, images with large objects and important object categories tend to be more specific. Number of annotated objects in an image does not correlate with specificity. Results are on the MEM-5S dataset.}
\label{fig:correlate_specificity}
\end{figure}

In the main paper, we described how \gls{automated specificity} correlated with \glslink{human specificity}{human-measured specificity}. Figure~\ref{fig:automated_vs_human} further illustrates this using a scatter plot. We also studied how various image properties correlated with specificity. In Figure~\ref{fig:correlate_specificity}, we illustrate these correlations via scatter plots. 

\section{Predicting specificity} \label{sec:predicted_specificity}
As we have shown, certain image-level objects and attributes make some images more specific than others. This means that specificity may be predictable using image features alone.

\begin{figure}[htb]
\begin{center}
   \includegraphics[width=0.8\linewidth]{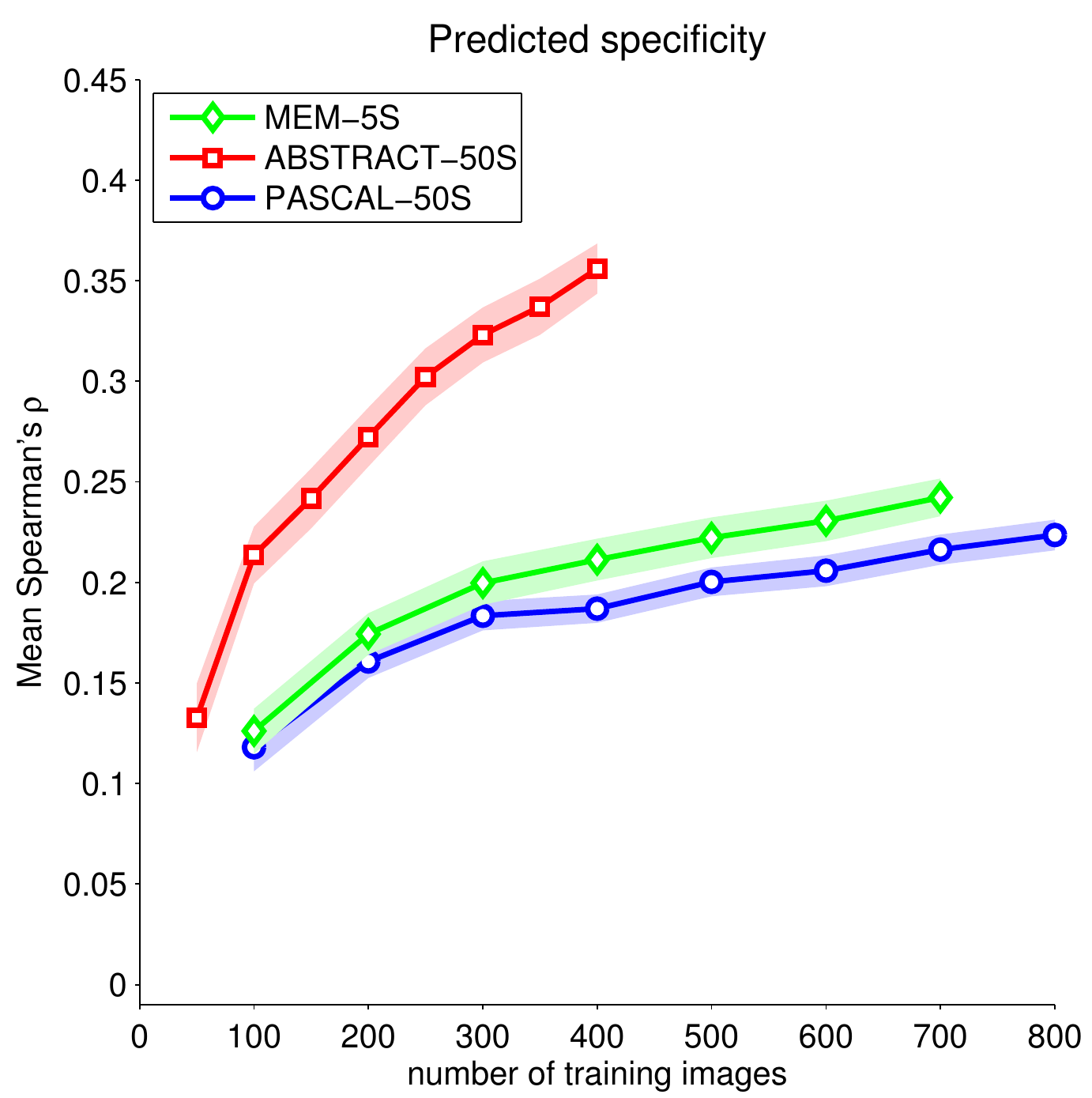}
\end{center}
   \caption{Spearman's rank correlation between \glslink{predicted specificity}{predicted} and \gls{automated specificity} for increasing number of training images (averaged across 50 random runs). \glslink{automated specificity}{Automated specificity} (Section 3.1.2 in main paper)  uses 5, 48 and 50 sentences per image for the three datasets, MEM-5S, ABSTRACT-50S and PASCAL-50S to estimate the specificity of the image. \glslink{predicted specificity}{Predicted specificity} (Section~\ref{sec:predicted_specificity}) uses only image features to predict the specificity. Different datasets have different number of images in them, hence they stop at different points on the x-axis. Higher correlation is better. The error bars represented by shaded colors show the standard error of the mean (SEM).}
\label{fig:predict_specificity}
\end{figure}

To test this, a $\nu$-SVR with an RBF kernel is trained on a randomly chosen subset of images represented by their DECAF-6 features~\cite{donahue2013decaf} in the MEM-5S and PASCAL-50S datasets. In the ABSTRACT-50S dataset, the image features are a concatenation of object occurrence, their absolute position, depth, flip angle, object co-occurrence, and clip art category~\cite{zitnick2013bringing}. For prediction, 188 images are set aside in the MEM-5S dataset, 200 images in the PASCAL-50S dataset, and 100 images in the ABSTRACT-50S dataset. Figure~\ref{fig:predict_specificity} shows that as the number of images used for training increases, the correlation of the \gls{predicted specificity} with the \glslink{automated specificity}{ground truth automated specificity} increases. We see that specificity can indeed be predicted from just image content better than chance. The use of semantic features (\eg occurence of objects) as opposed to low-level features (\eg DECAF-6) in the ABSTRACT-50S dataset seem to make it easier to predict specificity for that dataset as compared to the MEM-5S and PASCAL-50S datasets. Note that here we are directly predicting \gls{automated specificity} whereas in the main paper, we focused on predicting the two parameters of the Logistic Regression model. The latter is directly relevant to the image search application on which we demonstrated the benefit of specificity. 
\section{Detailed explanation of automated specificity computation}

\begin{figure}[htb]
\begin{center}
   \includegraphics[width=0.9\linewidth]{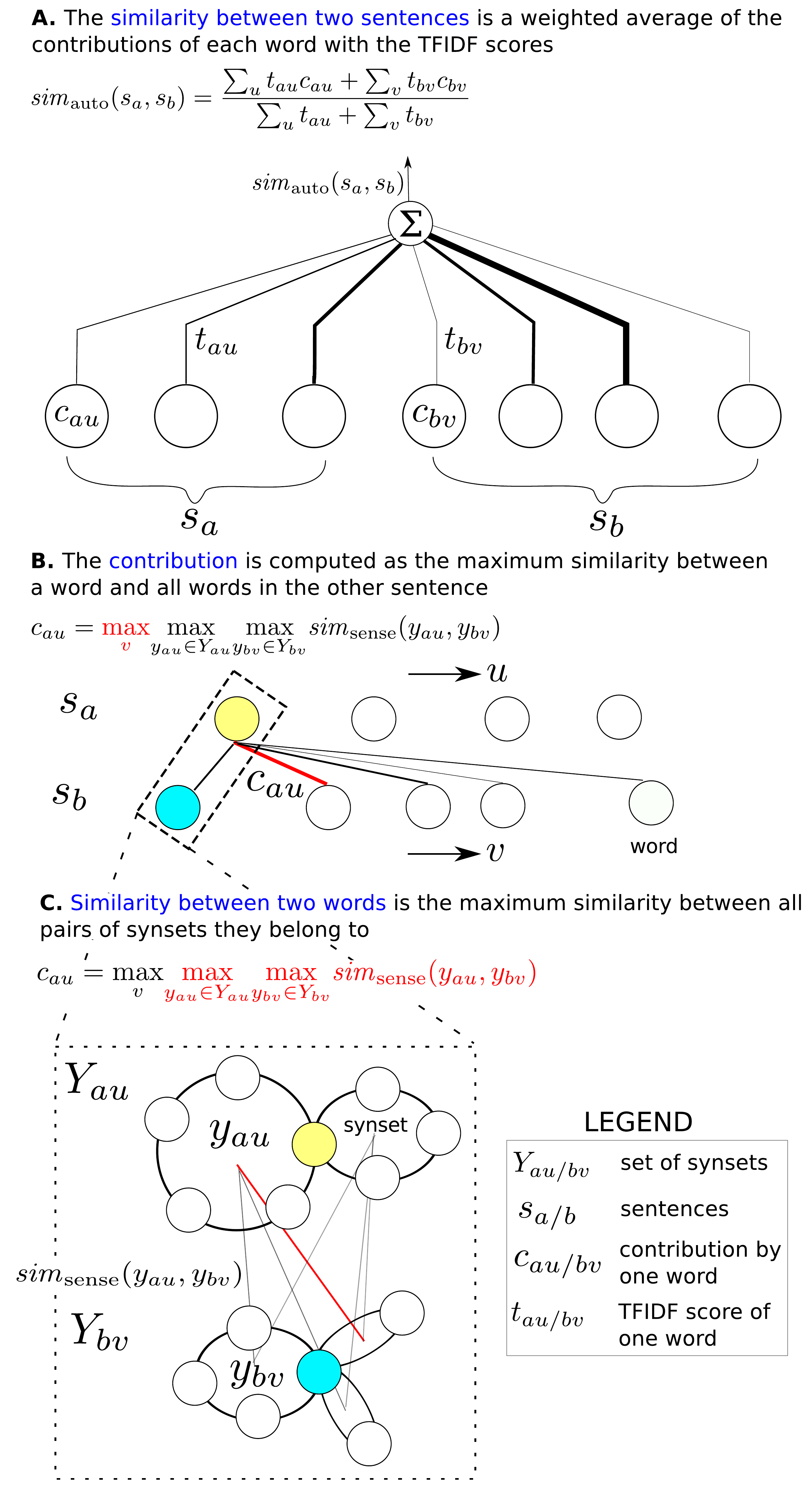}
\end{center}
   \caption{Illustration of our approach to compute automated sentence similarity.}
\label{fig:explain_similarity}
\end{figure}

\begin{figure*}[htb]
\begin{center}
   \includegraphics[width=1.0\linewidth]{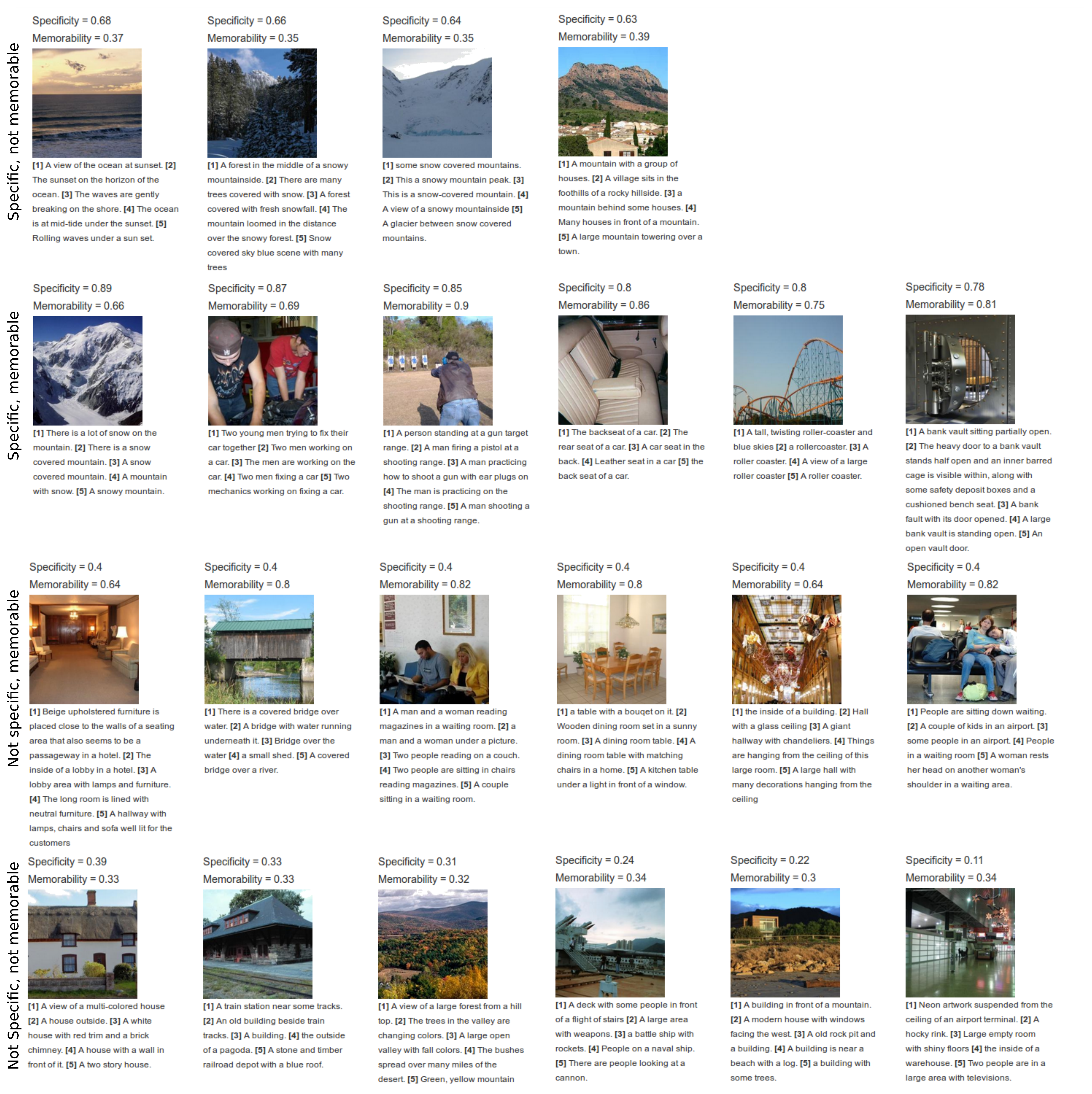}
\end{center}
   \caption{Examples illustrating the similarity and distinctions between image memorability~\cite{Isola2011} and image specificity.}
\label{fig:specificity_vs_memorability}
\end{figure*}

In Figure~\ref{fig:explain_similarity}, we visually illustrate the equations and notations used to automatically compute the similarity between two sentences (described in Section 3.1.2 in the main paper). To measure specificity automatically given the $N$ descriptions for image $i$, we first tokenize the sentences and only retain words of length three or more. This ensured that semantically irrelevant words, such as `a', `of', \etc, were not taken into account in the similarity computation (a standard stop word list could also be used instead). We identified the synsets (sets of synonyms that share a common meaning) to which each (tokenized) word belongs using the Natural Language Toolkit~\cite{bird2006nltk}. Words with multiple meanings can belong to more than one synset. Let $Y_{au} = \{y_{au}\}$ be the set of synsets associated with the $u\text{-th}$ word from sentence $s_a$. 

Every word in both sentences contributes to the automatically computed similarity $\mathit{sim}_\mathrm{auto}(s_a, s_b)$ between a pair of sentences $s_a$ and $s_b$. The contribution of the $u\text{-th}$ word from sentence $s_a$ to the similarity is $c_{au}$. This contribution is computed as the maximum similarity between this word, and all words in sentence $s_b$ (indexed by $v$) (Figure~\ref{fig:explain_similarity}B). The similarity between two words is the maximum similarity between all pairs of synsets (or senses) to which the two words have been assigned (Figure~\ref{fig:explain_similarity}C). We take the maximum because a word is usually used in only one of its senses. Concretely,

\vspace{-10pt}
\begin{equation}
c_{au} = \max_{v} \max_{y_{au} \in Y_{au}} \max_{y_{bv} \in Y_{bv}} \mathit{sim}_\mathrm{sense} (y_{au},y_{bv})
\end{equation}

The similarity between senses $\mathit{sim}_\mathrm{sense} (y_{au},y_{bv})$ is the shortest path similarity between the two senses on WordNet~\cite{miller1995wordnet}. We can similarly define $c_{bv}$ to be the contribution of $v\text{-th}$ word from sentence $s_b$ to the similarity $\mathit{sim}_\mathrm{auto}(s_a, s_b)$ between sentences $s_a$ and $s_b$.

The similarity between the two sentences is defined as the average contribution of all words in both sentences, weighted by the importance of each word (Figure~\ref{fig:explain_similarity}A). Let the importance of the $u\text{-th}$ word from sentence $s_a$ be $t_{au}$. This importance is computed using term frequency-inverse document frequency (TF-IDF) using the scikit-learn software package~\cite{pedregosa2011scikit}. Words that are rare in the corpus but occur frequently in a sentence contribute more to the similarity of that sentence with other sentences. So we have

\vspace{-10pt}
\begin{equation} \label{eq:automated_sent_sim}
\mathit{sim}_\mathrm{auto}(s_a, s_b) = \frac{\sum_{u} t_{au}c_{au} + \sum_{v} t_{bv}c_{bv}}{\sum_{u} t_{au} + \sum_{v} t_{bv}}
\end{equation}

The denominator in Equation~\ref{eq:automated_sent_sim} ensures that the similarity between two sentences is independent of sentence-length and is always between 0 and 1.

\section{Specificity vs. Memorability}

\begin{figure*}[htb]
\begin{center}
   \includegraphics[width=0.8\linewidth]{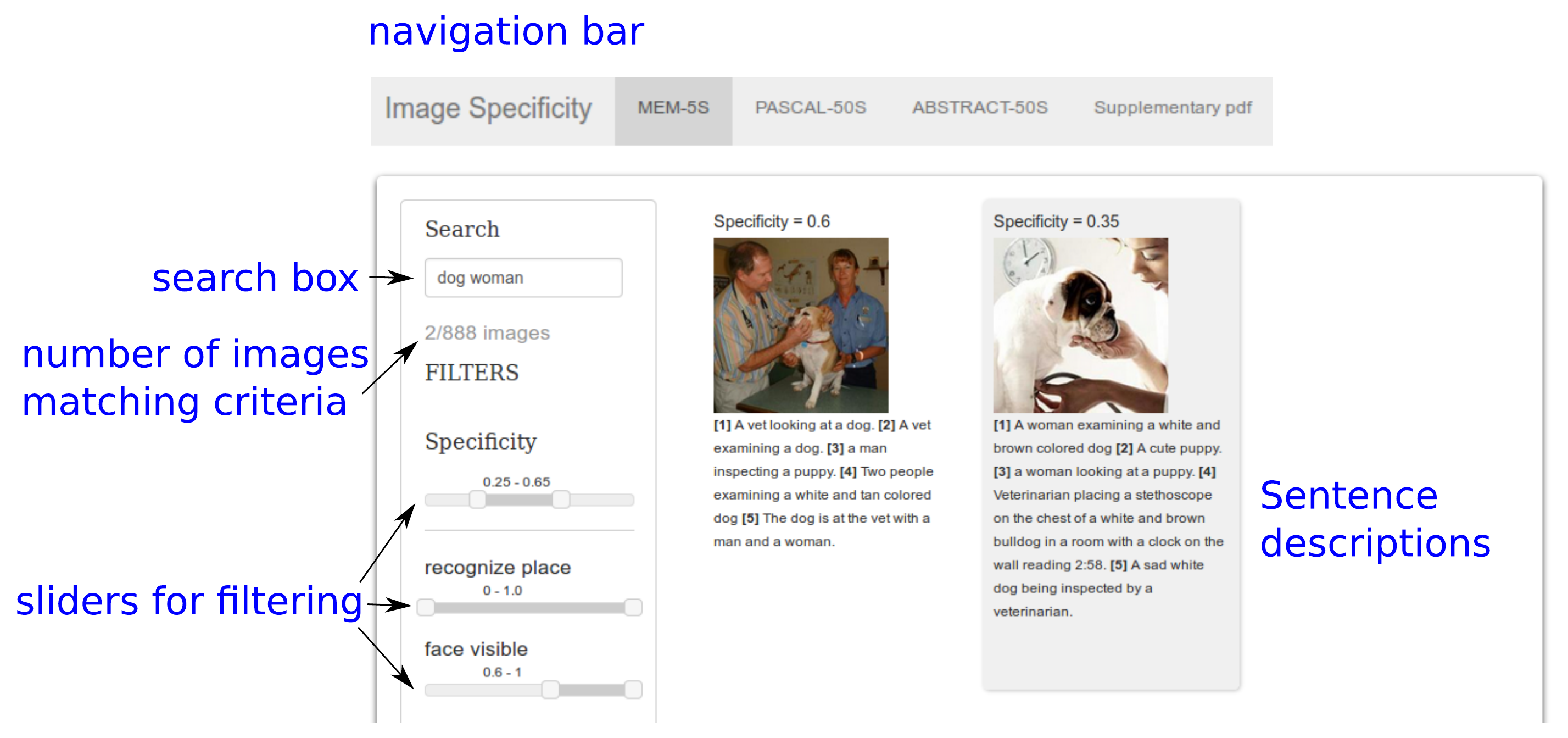}
\end{center}
   \caption{Dataset browser for exploring the datasets. Available on the authors' webpages.}
\label{fig:search_function}
\end{figure*}

In our paper, we have shown that specificity and memorability are correlated. However, they are distinct concepts and measure different properties of the image. In particular, we have shown that peaceful and picture-perfect scenes are negatively correlated with memorability but have no effect on specificity. In Figure~\ref{fig:specificity_vs_memorability}, we show examples of images that are specific/not specific and memorable/not memorable. Note how outdoor scenes tend to be not very memorable but can have a reasonably high specificity score.

\section{Website for exploring datasets}

Here, we describe the website interface available on the authors' webpages that can be used to explore the datasets used in the paper. A navigation bar on top of the website allows users to switch between different datasets. Figure~\ref{fig:search_function} shows how the search function can be used to look for sentences containing the words ``dog" and ``woman". Up to a maximum of 6 words can be added in the search box. Only whole words are matched. The reader should note that the website does not implement the text-based search algorithms discussed in the paper. It is meant for only browsing the datasets. Sliders on the left allow the user to filter images according to a range of scores that the images satisfy. All the criteria are combined using logical AND to display the filtered images. The number of images matching the search criteria gives the user an idea of how often two or more criteria are satisfied concurrently. The benefit of using such a website is that it can give the readers an intuition of the underlying data and factors that affect specificity. We have added sliders for the attributes that correlate most (top 10) and least (bottom 10) with specificity (for the MEM-5S dataset). It is also possible to filter by average length of the sentences and the memorability score.